\documentclass[english]{article}
\usepackage[T1]{fontenc}
\usepackage[latin9]{inputenc}
\usepackage[letterpaper]{geometry}
\usepackage{amstext}
\usepackage{amssymb,amsfonts,amsmath}
\usepackage{epsfig}
\usepackage{amsthm}
\usepackage{amssymb}
\usepackage{amsmath}
\usepackage{graphics}
\usepackage{graphicx}
\usepackage{dcolumn}
\usepackage{bm}
\usepackage{subfigure}
\usepackage{xcolor}
\usepackage{color}
\usepackage{babel}
\usepackage{setspace}
\doublespace

\makeatletter
\newcommand{\lyxaddress}[1]{
\par {\raggedright #1
\vspace{1.4em}
\noindent\par}
}

\makeatother

\begin{document}

\title{A Boosting Method to Face Image Super-resolution}

\date{}

\author{Shanjun Mao$^{\text{}1}$, Da Zhou$^{\text{}1}$, Yiping Zhang$^{\text{}2}$, Zhihong Zhang$^{\text{}2}$, Jingjing Cao$^{\text{}3}$}
\maketitle
\lyxaddress
{1. School of Mathematical Sciences, Xiamen University, Xiamen 361005, PR China\\
2. Software School, Xiamen University, Xiamen 361005, PR China\\
3. School of Logistics Engineering, Wuhan University of Technology, Wuhan 630047, PR China
}

\begin{abstract}
Recently sparse representation has gained great success in face image super-resolution. The conventional sparsity-based methods enforce sparse coding on face image patches and the representation fidelity is measured by $\ell_{2}$-norm. Such a sparse coding model regularizes all facial patches equally, which however ignores distinct natures of different facial patches for image reconstruction. In this paper, we propose a new weighted-patch super-resolution method based on AdaBoost. Specifically, in each iteration of the AdaBoost operation, each facial patch is weighted automatically according to the performance of the model on it, so as to highlight those patches that are more critical for improving the reconstruction power in next step. In this way, through the AdaBoost training procedure, we can focus more on the patches (face regions) with richer information.  Various experimental results on standard face database show that our proposed method outperforms state-of-the-art methods in terms of both objective metrics and visual quality.
\end{abstract}

\section{Introduction}
A fundamental challenge in practical face recognition system is to increase the resolution of blurry face images. The low-resolution (LR for short) face images not only bring down the human visual experience but also adversely affect the performance of the followed face recognition and analysis. To alleviate this problem, image super-resolution (SR) attempts to reconstruct the original high-resolution (HR) image $\mathbf{x}$ from its degraded observed version $\mathbf{y}$.  It can be generally formulated as:

\begin{equation}
\mathbf{y} = \mathbf{SHx}+ e~,
\label{model1}
\end{equation}

here we assume that a LR image $\mathbf{y}$ is generated from a HR image $\mathbf{x}$ by first convolving $\mathbf{x}$ with a low pass filter $\mathbf{H}$, to reduce aliasing, and then down-sampling to the desired size with $\mathbf{S}$. The dimension of $\mathbf{y}$ is significantly smaller than that of $\mathbf{x}$; thus there exist multiple corresponding HR images $\mathbf{x}$ for a specific LR image $\mathbf{y}$. To cope with this ill-posed nature of image restoration, regularization techniques based on a priori knowledge are required, and it can be formulated as the following minimization problem:

\begin{equation}
\min_{\mathbf{x}}\frac{1}{2}\|\mathbf{SHx}-\mathbf{y}\|_{2}^{2}+\lambda R(\mathbf{x}),
\label{model2}
\end{equation}

where $\frac{1}{2}\|\mathbf{SHx}-\mathbf{y}\|_{2}^{2}$ is the $\ell_{2}$ data-fidelity term, $R(\mathbf{x})$ is called the regularization term denoting image prior and $\lambda$ is the regularization parameter.

Due to that the image prior knowledge is imperative and of great important to eliminate the uncertainty of image recovery, various priors have been designed, ranging from half quadrature formulation \cite{geman1992constrained}, Mumford-Shah (MS) model \cite{mumford1989optimal}, and total variation (TV) models \cite{rudin1992nonlinear}\cite{chambolle2004algorithm}. These regularization terms demonstrate high effectiveness in preserving edges and recovering smooth regions. However, they usually smear out image details and cannot deal well with fine structures.

Due to the success of sparse representation used in incomplete signal recovering, a series of methods based on sparse representation are developed for face image super-resolution. These methods assume that each patch from the images considered can be well represented using a linear combination of few atoms from a dictionary. By forcing LR patch and the corresponding HR patch to have the same sparse coefficient, Yang et al. \cite{yang2010image} are the first to introduce the idea of sparse representation to the face image SR. The method offline trains a HR and LR dictionary to sparsely decompose HR and LR image patches, respectively. Given a LR patch input, a sparse coefficient vector is computed using the LR dictionary by solving an $\ell_{1}$-norm minimization problem. The desired HR patch is reconstructed by combining the HR dictionary. The similar intuitive is used in \cite{chang2010face}. Chang et al. \cite{chang2010face} used coupled over-complete dictionaries and sparse representation to synthesize face sketch which obtained better result. In order to fully use the structure information of facial images, ELad et al. \cite{aharonk} used sparse representation for photo-ID image compressing by adapting to the image content. The prior of face position can be incorporated into face super-resolution. Ma et al. \cite{ma2012sparse} took face position information as a feature and proposed a position-patch based face hallucination method. It estimated a HR image patch using the image patches at the same positions of all training face image. Specifically, the coding coefficients estimated via constrained least square (CLS) in each face region are used to generate the HR patch of the corresponding position. However, when the number of the training images is much larger than the dimension of the patch, the CLS problem is underdetermined and the solution is not unique. To address the biased solution problem caused by least square estimation, Jung et al. \cite{jung2011position} provided a position-patch based face hallucination method using convex optimization, which obtained the optimal weights for face hallucination and achieved a better results than Ma's \cite{ma2012sparse}. However, this sparse coding based method \cite{jung2011position} fails to consider the manifold geometric structure of the face data that is important for image representation and analysis. Zhang et al. \cite{zhang2012image} presented a dual-dictionary learning method to recover more image details, in which both the main and the residual dictionaries are learned by sparse representation. These SR based methods give impressive improvements for experimental noise free faces. However, due to the under sparse nature of noisy images, they usually perform unsatisfactorily in the presence of noise.

The above sparse representation based super-resolution approaches are based on the minimization of mean-squared-error (MSE) between the input LR image patches and the reconstructed SR image patches (i.e., $\|\mathbf{y} - \mathbf{D}\alpha\|_{2}^{2}$). In fact, the fidelity term has a high impact on the final coding results because it ensures that the given signal $\mathbf{y}$ can be faithfully represented by the dictionary $\mathbf{D}$. From the view of maximum likelihood estimation (MLE), defining the fidelity term with $\ell_{2}$-norm actually assumes that the coding residual $e = \mathbf{y} - \mathbf{D}\alpha$ follows Gaussian distribution. Such coding fidelity treats all the face image patches equally, and it does not differentiate the natures of different facial patches, where the facial patches in the different regions (patch positions) of human face may have distinct contribution to face image reconstruction. Intuitively, face regions (such as mouth, eyes, nose) are rich in texture containing more high-frequency details; thus, facial patches in this regions are expected to be assigned with high weight values to ensure very
small residuals.

To improve the robustness and effectiveness of face hallucination, we propose a weighted-patch method based on AdaBoost \cite{freund1997decision}.
Instead of regularizing all facial patches equally, our proposed method discriminates different patches via an AdaBoost training procedure.
In each step of the AdaBoost operation, each patch is re-weighted automatically according to the performance of the learner on it. The larger the error is, the more the patch is weighted in next step. In this way, the AdaBoost process can make us focus more on the facial regions that are more difficult to be correctly captured. Note that the difficult patches are quite in line with the ones with richer information, our method just provides an effective way to screen out the patches that are more critical for improving the power of the image reconstruction.

\section{Related Work}

We review some of the related previous works in this section, which will lay the foundation for the derivation of our approach later.

\subsection{Super-resolution via Couple Dictionaries and Sparse Coding}

Yang et al. \cite{yang2010image} proposed an approach for super-resolution based on sparse representation. Given $\mathbf{D_{l}} \in \Re^{M \times K}$ be an over-complete LR dictionary of $K$ prototype signal-atoms, $\mathbf{D_{h}} \in \Re^{N \times K}$ be the corresponding over-complete HR dictionary of $K$ prototype signal-atoms, where $N$ and $M$ are the dimensions of a HR image patch and LR image patch, respectively. Yang et al. \cite{yang2010image} start from a large collection of low resolution (LR) and high resolution (HR) training patch pairs and use a sparsity constraint to jointly train the LR and HR dictionaries by assuming that LR patches and their corresponding HR counterparts shares the same sparse coding vector. The optimal dictionary pair $\{\mathbf{D_{h}},\mathbf{D_{l}}\}$ is obtained by minimizing

\begin{equation}
\min_{\mathbf{D_{h}},\mathbf{D_{l}},Z}\frac{1}{N}\|C_{h}-\mathbf{D_{h}}Z\|_{2}^{2}+ \frac{1}{M}\|A_{l}-\mathbf{D_{l}}Z\|_{2}^{2}+\lambda(\frac{1}{N}+\frac{1}{M})\|Z\|_{1}~,
\label{yang2}
\end{equation}

Once the dictionaries are trained, the input LR image is divided into overlapped patches, and each patch $\mathbf{A}$ can be sparsely encoded by a learned LR dictionary  $\mathbf{D_{l}}$ using the following formulation:

\begin{equation}
\min_{\mathbf{\alpha}}\|F\mathbf{D_{l}}\mathbf{\alpha} - F\mathbf{y}\|_{2}^{2}+\lambda\|\mathbf{\alpha}\|_{1}~,
\label{yang1}
\end{equation}

where $F$ is a feature extraction operator, $\mathbf{\alpha}$ is the sparse representation and $\lambda$ is a weighting factor. The corresponding HR patch is reconstructed by  $\mathbf{D_{h}}$ and $\mathbf{\alpha}$ with $\mathbf{D_{h}\alpha}$. Finally, the HR image can be obtained by aggregating all the estimated HR patches into a whole image. One problem of Yang's work \cite{yang2010image} is that the dictionary training process is time-consuming. Therefore, it will be much efficient if we can project the patch vectors into a lower subspace while preserving most of their average energy.

Zeyde et al. \cite{zeyde2010single} improved the work of Yang et al. \cite{yang2010image} with less computation time and better estimation result. They perform dimensionality reduction of LR image patches via Principal Component Analysis (PCA) to improve the execution speed. With the training patch pairs $\{\mathbf{C_{h}},\mathbf{A_{l}}\}$ prepared, they firstly learn the LR dictionary as:

\begin{equation}
\mathbf{\alpha} = \arg\min_{\mathbf{\alpha}}\|\mathbf{A_{l}}-\mathbf{D_{l}}\mathbf{\alpha} \|_{2}^{2}+\lambda\|\mathbf{\alpha}\|_{1}~,
\label{zedyde1}
\end{equation}

The above optimization formula can be solved by K-SVD \cite{aharon2006img} and Orthogonal Matching Pursuit \cite{aharon2006img}. By the same assumption with Yang's work \cite{yang2010image}, the sparse code $\mathbf{\alpha}$ trained from above can be utilized in constructing the HR dictionary $\mathbf{D_{h}}$. And the $\mathbf{D_{h}}$ training can be formulated as a least square regression problem:

\begin{equation}
\mathbf{D_{h}} = \min_{\mathbf{\alpha}}\|\mathbf{C_{h}}-\mathbf{D_{h}}\mathbf{\alpha} \|_{2}^{2}~,
\label{zedyde2}
\end{equation}

Hence, a straightforward least-square solution of $\mathbf{D_{h}} $ can be obtained by:

\begin{equation}
\mathbf{D_{h}} = \mathbf{C_{h}}\mathbf{\alpha^{T}}(\mathbf{\alpha}\mathbf{\alpha^{T}})^{-1}~,
\label{zedyde3}
\end{equation}

From above, it is clear that this method can only train for LR dictionary and its corresponding sparse code, leading to more time saving and less computation complexity.
Despite the improvements, the use of OMP during sparse coding is clearly the bottleneck.


\subsection{Anchored Neighborhood Regression}
Starting from the same dictionaries training by K-SVD with OMP algorithms in \cite{zeyde2010single}, the Anchored Neighborhood Regression (ANR) approach \cite{timofte2013anchored} proposes to relax the sparsity constraint in  Eq. \ref{yang1} and  reformulates the patch representation problem as a least squares (LS) $\ell_{2}$-norm regression. The method uses the local neighborhoods of dictionary (i.e. $\mathbf{N_{l}}$ and $\mathbf{N_{h}}$) with a specific size instead of the entire dictionary used in \cite{yang2010image}. Compared with solving $\ell_{1}$-norm minimization  which is computationally demanding, the $\ell_{2}$-norm regression turns the problem into  Ridge Regression \cite{timofte2014adaptive} and a closed-form solution can be obtained.

\begin{equation}
\min_{\mathbf{\beta}}\|\mathbf{A} - \mathbf{N_{l}}\mathbf{\beta}\|_{2}^{2}+\lambda\|\mathbf{\beta}\|_{2}~,
\label{anchored1}
\end{equation}

where $\mathbf{N_{l}}$ is the LR neighborhood of input patch $\mathbf{A}$ chosen from $\mathbf{D_{l}}$. The algebraic solution of the coefficient vector $\mathbf{\beta} $ can be written as:

\begin{equation}
\mathbf{\beta} = (\mathbf{N_{l}^{T}}\mathbf{N_{l}}+\lambda\mathbf{I})^{-1}\mathbf{N_{l}^{T}}\mathbf{A}~,
\label{anchoredcoeficient}
\end{equation}

the coefficients of $\mathbf{\beta} $ are then applied to the corresponding HR neighborhood  $\mathbf{N_{h}}$ to reconstruct the HR patch $\mathbf{x}$

 \begin{equation}
\mathbf{x} = \mathbf{N_{h}}(\mathbf{N_{l}^{T}}\mathbf{N_{l}}+\lambda\mathbf{I})^{-1}\mathbf{N_{l}^{T}}\mathbf{A} = \mathbf{P} \mathbf{A}~,
 \label{anchored2}
 \end{equation}

where $\mathbf{P}$ is the projection matrix for dictionary atom $\mathbf{d_{l}}$. Given the trained couple dictionaries, for each LR dictionary atom $\mathbf{d_{l}}$, we search for its $K$ nearest neighborhoods of dictionary $\mathbf{N_{l}}$ by correlation between the whole dictionary atoms. Then, based on the neighborhoods of $\mathbf{d_{l}}$, a separate projection matrix $\mathbf{P}$ can be computed. Therefore, the projection matrix can be obtained offline and the procedure of SR for ANR at test time becomes mainly a nearest neighbor search followed by a matrix multiplication for each input patch.

Although the effectiveness of sparse representation has been proven, the spatial information is lost during the coding phase. We believe that the amount of information in different face regions is different and the spatial information should also be included in the face  image reconstruction.


\section{The Proposed Algorithm}
\subsection{AdaBoost Procedure}
In a wide variety of classification and regression problems, boosting techniques have proven to be an effective method for reducing bias and variance, improving misclassification rates and regression effects. In this paper, we mainly adopt the adaptive boosting(AdaBoost) algorithm.

$\bold{AdaBoost \ for \ Classification}$ \ \ \ \ In binary classification, let $\bold{T} = \{(x_1, y_1), (x_2, y_2), \cdots, (x_N, y_N)\}$ be the input data, where the $i$-th input point $x_i \in \mathbb{X} \subseteq \mathbb{R}^d$, and its corespoding label $y_i \in \mathbb{Y} = \{-1, 1\}$. AdaBoost uses the next algorithm to learn a series of weak-classifiers, then combines these weak-classifiers to get a strong-classifier \cite{freund1997decision}.\\
AdaBoost proceeds as follows.

$\bold{Input}$:  training set $\bold{T} = \{(x_1, y_1), (x_2, y_2), \cdots, (x_N, y_N)\}$, where $x_i \in \mathbb{R}^d, \ \ y_i \in \{-1, 1\}$; weak-learning algorithm;

$\bold{Output}$:  the final classifier $G(x)$.

$1_{\cdot}$ \ \ \ Initialize the weights of each input points

\begin{center}
$\bold{W}_1 = (w_{11}, w_{12}, \cdots, w_{1N}), \ w_{1i} = \frac{1}{N}, \ i = 1,2, \cdots, N$
\end{center}

$2_{\cdot}$ \ \ \ For $m = 1,2, \cdots, M$

 \ $a_{\cdot}$ \ \ Using the weight $\bold{W}_m$, learn weak-classifier

 \begin{center}
 $G_m(x): \mathbb{X} \rightarrow \{-1, 1 \}$
 \end{center}

 \ $b_{\cdot}$ \ \ Compute the error rate for $G_m(x)$ on the training set
 \begin{equation}
 e_m = P(G_m(x_i) \neq y_i) = \sum \limits_{i=1}^{N} w_{m,i} I(G_m(x_i) \neq y_i)
 \end{equation}

 \ $c_{\cdot}$ \ \ Compute the coefficient of $G_m(x)$
 \begin{equation}
 \beta_m = \frac{1}{2} \log \frac{1 - e_m}{e_m}
 \end{equation}

 \ $d_{\cdot}$ Update the weights of each input points
 \begin{equation}
 \bold{W}_{m+1} = (w_{m+1,1}, \cdots, w_{m+1,i}, \cdots, w_{m+1,N})
 \end{equation}
 \begin{equation}
 w_{m+1,i} = \frac {w_{m,i}}{Z_m} \exp {(- \beta_m y_i G_m(x_i))}, \ \ i = 1,2, \cdots, N
 \end{equation}
 where $Z_m$ is a normalization constant
 \begin{equation}
 Z_m = \sum \limits_{i=1}^N w_{m,i} \exp (-\beta_m y_i G_m(x_i))
 \end{equation}

 $3_{\cdot}$ \ \ \ Construct the linear combination of these weak-classifiers
 \begin{equation}
 f(x) = \sum \limits_{m=1}^M \beta_m G_m(x)
 \end{equation}
 and get the final classifier
 \begin{equation}
 G(x) = \mathrm{sign} (f(x)) = \mathrm{sign} \bigg ( \sum \limits_{m=1}^M \beta_m G_M(x) \bigg )
 \end{equation}

 $\bold{AdaBoost \ for \ Regression}$ \ \ \ \ If boosting's effectiveness extends beyond classification problems then we might expect that the boosting of simplistic regression models could result in a richer class of regression models. So, during the process of AdaBoost for classification, we adjust some steps to make the above algorithm adapt the regression task \cite{drucker1997improving}. Training set $\bold{T} = \{(x_1, y_1), (x_2, y_2), \cdots, (x_N, y_N)\}$, where $x_i \in \mathbb{R}^d, \ \ y_i \in \mathbb{R}$.\\
 Similar to the AdaBoost algorithm for classification we set
 \begin{equation}
 e_m = \sum \limits_{i=1}^N w_{m,i} L \big ( |y_i - F_m (x_i)| \big )
 \end{equation}
 where $F_m(x): \mathbb{X} \rightarrow \mathbb{R}$. The loss $L$ may be of any function form as long as $L \in [0, 1]$, and we have three candidate loss functions:
 \begin{equation}
 L \big ( |y_i - F_m (x_i)| \big ) = \frac{|y_i - F_m (x_i)|}{\max |y_i - F_m (x_i)|} \ \ \ (linear)
 \end{equation}
 \begin{equation}
 L \big ( |y_i - F_m (x_i)| \big ) = \bigg ( \frac{|y_i - F_m (x_i)|}{\max |y_i - F_m (x_i)|} \bigg )^2 \ \ \ (square law)
 \end{equation}
 \begin{equation}
 L \big ( |y_i - F_m (x_i)| \big ) = 1 - \exp \bigg [ - \frac{|y_i - F_m (x_i)|}{\max |y_i - F_m (x_i)|} \bigg ] \ \ \ (exponential)
 \end{equation}
 and
 \begin{equation}
 w_{m+1,i} = \frac {w_{m,i}}{Z_m} \big ( \frac {e_m}{1 - e_m} \big )^{1 - L \big ( |y_i - F_m (x_i)| \big ) }
 \end{equation}
 where $Z_m$ is a normalization constant
 \begin{equation}
 Z_m = \sum \limits_{i=1}^N w_{m,i} \big ( \frac {e_m}{1 - e_m} \big )^{1 - L \big ( |y_i - F_m (x_i)| \big ) }
\end{equation}

\subsection{Weighted-patch algorithm via AdaBoost}

In sparse coding, for each low-resolution patches $\bold{A}_i$, we use the equatz to obtain its corresponding sparse representation, and we write these equations together:
\begin{equation}
\bm{\alpha} = \arg \min_{\bm{\alpha}} \ \big(\sum_{i = 1}^N||\bold{A}_i - \bold{D}_l\alpha_i||_2^2 + \lambda||\alpha_i||_1\big)
\end{equation}
where $\bm{\alpha} = [\alpha_1,\cdots, \alpha_N]^T$.\\
In the above equation, each patch is been treated equally, but we have already know that some parts on the face, such as eyes and nose, are more important than the rest. So next, we consider adding weight for each patch to reflect patches' importance.

$\bold{The \ Procedures}$ \ \ \ \ Each face image $\bold{A}$ is divided into overlapped patches. Each patch within $\bold{A}$ is mapped to a vector, and the vector is regarded as the feature of the patch, denoted by $\{\bold{A}_i\}_{i = 1}^N$ where $N$ is the number of patches within $\bold{A}$.

$\bold{Input}$: training dictionaries $\bold{D}_h$ and $\bold{D}_l$, a low-resolution image $\bold{X}$

$\bold{Output}$: super-resolution image $\bold{Y}^{\ast}$

$1_{\cdot}$ \ \ \ Choose P high-resolution images, which are not used to learn the dictionaries, and get their blurred and downsampled versions to form the training set $\mathbb{S} \in R^{d_L \ast P}$;

$2_{\cdot}$ \ \ \ Initialize the weights of each patch
\begin{equation}
\bold{W_1} = (w_{11}, w_{12}, \cdots, w_{1N}), \ w_{1i} = \frac{1}{N}, \ i = 1, 2, \cdots, N
\end{equation}

$3_{\cdot}$ \ \ \ For $m = 1, 2, \cdots, M$

 $a_{\cdot}$ \ \ For each image in the train set $\mathbb{S}$, $\bold{A} \in \mathbb{R}^{d_L \ast 1}$, using the weight $\bold{W}_m$, get the corresponding sparse representations
\begin{equation}
\begin{split}
\min _{\bm{(\alpha_{m,1}, \alpha_{m,2}, \cdots, \alpha_{m,N})}} & \bigg ( \sum \limits_{i=1}^N \big ( \| e^{w_{m,i}} (\bold{A}_i - \bold{D}_l \bm{\alpha_{m,i}}) \|_2^2 + \lambda \| \bm{\alpha_{m,i}} \|_1 \big ) \\
 & + \theta \| \bold{B}_m - \bold{C} \|_2^2 \bigg )
\end{split}
\end{equation}
where $A_i \in \mathbb{R}^{d_l \ast 1}$ is one of the patches within $\bold{A}$, $\bm{\alpha}_{m,i} \in \mathbb{R}^{k \ast 1}$ is its corresponding sparse representation, $\bold{C} \in \mathbb{R}^{d_H \ast 1}$ is $\bold{A}$'s original high-resolution image and $\bold{B}_m$ is $\bold{A}$'s resolved high-resolution image in the $m$-th step constructed by the high-resolution patches $\bold{B}_{m,i} = \bold{D}_h \bm{\alpha}_{m,i}$, in other words, $\bold{B}_{m,i}, \ i = 1, 2, \cdots, N$ are the patches within $\bold{B}_m$, $w_{m,i}$ is the weight of $i$-th patch in the $m$-th step.\\
And Giving the $\bold{W}_m$, the hyper-resolution equation in the $m$-th, $F_m (\bold{X})$, is:
\begin{equation}
\min _{(\bm{\alpha}_{m,1}, \bm{\alpha}_{m,2}, \cdots, \bm{\alpha}_{m,N})} \bigg ( \sum \limits_{i=1}^N \big ( \| e^{w_{m,i}} (\bold{X}_i - \bold{D}_l \bm{\alpha}_{m,i}) \|_2^2 + \lambda \| \bm{\alpha}_{m,i} \|_1 \big ) \bigg )
\end{equation}
where $\{ \bold{X}_i \}_{i=1}^N$ are the patches with the testing low-resolution picture $\bold{X}$ and put the patches $\{ \bold{D}_h \alpha_{m,i} \} _{i=1}^N$ together to form the high-resolution image $F_m (\bold{X})$.

 $b_{\cdot}$ \ \ Compute the error rate for $F_m (\bold{X})$ on the training image $\bold{A}$
 \begin{equation}
 e_m' = \sum \limits_{i=1}^N w_{m,i} L \big ( \| \bold{B}_{m,i} - \bold{C}_i \| \big )
 \end{equation}
 where $\{ \bold{C}_i \}_{i=1}^N$ are the patches within $\bold{C}$, the original high-resolution image of $\bold{A}$, $L$ is the loss function.\\
 Remark: we have P images in the training set $\mathbb{S}$, so, there should be P $e_m'$ on $\mathbb{S}$ for each training image. In this case, we choose the mean value of $e_m'$ as the final error rate of $F_m (\bold{X})$, recorded as $e_m = \rm{mean} (e_m')$.

 $c_{\cdot}$ \ \ Compute the coefficient of $F_m (\bold{X})$
 \begin{equation}
 \beta_m = \frac{1}{2} \log \frac{1 - e_m}{e_m}
 \end{equation}

 $d_{\cdot}$ \ \ Update the weights of patches
 \begin{equation}
 \bold{W}_{m+1} = (w_{m+1,1}, w_{m+1,2}, \cdots, w_{m+1,N})
 \end{equation}
 \begin{equation}
 w_{m+1,i} = \frac{w_{m,i}}{Z_m} \bigg ( \frac{e_m}{1 - e_m} \bigg )^{1 - L ( \| \bold{B}_{m,i} - \bold{C}_i \| )}
 \end{equation}
 where $Z_m$ is a normalization constant
 \begin{equation}
 Z_m = \sum \limits_{i=1}^N w_{m,i} \bigg ( \frac{e_m}{1 - e_m} \bigg )^{1 - L ( \| \bold{B}_{m,i} - \bold{C}_i \| )}
 \end{equation}

 $4_{\cdot}$ \ \ \ Construct the linear combination of $\{ F_m (\bold{X}) \}_{m=1}^M$
 \begin{equation}
 F (\bold{X}) = \sum \limits_{m=1}^M \frac{\beta_m}{\rm{sum}(\beta_m)} F_m (\bold{X})
 \end{equation}

 $5_{\cdot}$ \ \ \ Using gradient descent, find the closest image to $F (\bold{X})$ which satisfies the reconstruction constraint
 \begin{equation}
 \bold{Y}^{\ast} = \arg \min _{\bold{Y}} \| SH \bold{Y} - \bold{X} \|_2^2 + c \| \bold{Y} - F(\bold{X}) \|_2^2
 \end{equation}

\section{Experiment}
In this section, we conduct several experiments to evaluate the effectiveness of the proposed method, in terms of both objective metrics and visual quality. We compare the SR estimation results with several classical as well as state of the art SR methods including Bicubic interpolation, the sparse coding algorithms of Yang et al. \cite{yang2010image} and Zeyde et al. \cite{zeyde2010single}, Anchored Neighborhood Regression (ANR) approach \cite{timofte2013anchored}. These methods were configured using the same patch size and overlap as indicated below and configured using the optimal parameters provided in their respective papers. The face super-resolution performance is quantified by the Peak Signal to Noise Ratio (PSNR) \cite{wang2004image} between the ground truth face images and the super-resolved ones.

\subsection{Experimental configurations}

\textbf{Database:} The experiments conducted in this paper use different publicly available face datasets: i) FERET \cite{phillips2000feret}, ii) AR \cite{martinez1998ar}. All these face images were aligned by an automatic alignment algorithm using the eye positions, and then cropped to the size of 64 $\times$ 64 pixels. The LR images are formed by blurred and down-sampling (by a factor of 4 resulting the size of LR face images to be 16 $\times$ 16 pixels) the corresponding HR images. The 450 face images from FERET were used as HR dictionary training images and LR dictionary training images. Here, we adopted the same way of dictionary learning procedure as Zeyde et al. \cite{zeyde2010single}, which combining GOMP and K-SVD to train the dictionary pair. Then, we randomly select 50 face images from AR dataset as the testing image sets.

\textbf{Parameter Setting:} Empirically, we set the size as 16 $\times$ 16 pixels for HR patch and the overlap between neighbor patches as 4 pixels. The corresponding LR patch size is set to 4 $\times$ 4 with overlap of one pixel. After learning the dictionaries for high-resolution and low-resolution image patches, we can group the dictionary atoms into neighborhoods. Specifically, for each atom in the dictionary we find its K nearest neighbors based on the correlation between the dictionary atoms, which will represent its neighborhood. In our experiments, we set K to 40 and parameter $\lambda$ to 0.0001.

\subsection{Comparison experiments on benchmark database}

In this section, we perform experiments on the AR face dataset to demonstrate the effectiveness of the proposed method. Since the ground truth HR face images are available, we compare not only the visual quality but also the quantitative results of the reconstructed face images.

\begin{figure}[!t]
 	\centering
 	\includegraphics[scale=.45]{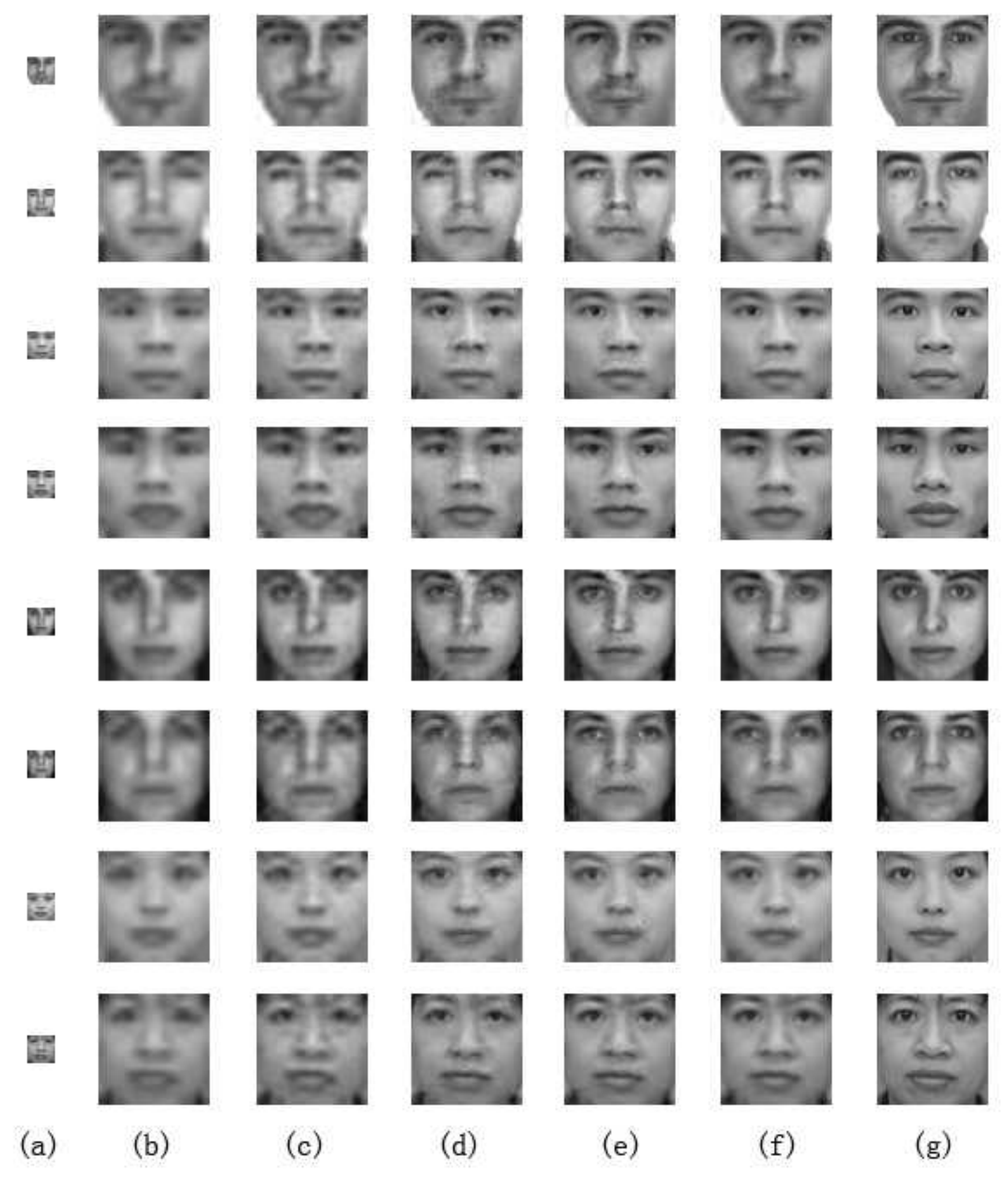}
 	\caption{{\bf Examples of our algorithm compared to other methods.}
	From left to right columns: (a) low-resolution input; (b) Bicubic interpolation; (c) Yang's method; (d) Zeyde's method; (e) ANR; (f) our proposed method; (g) original}
 	\label{visual}
 \end{figure}

In Fig. \ref{visual}, some reconstructed face images by different methods are compared.  It can be seen that the reconstructed face images by Bicubic interpolation is very blurry. Yang's results lost too many details and have many jaggy artifacts. The textures of ANR's results are heavily smoothed. Our method achieves better visual quality with fine facial details and better textures. Apparently, the reconstructed face images by our method are closest to the ground truth HR images. Besides, our method achieves higher PSNR values. Table 1 and 2 show the quantitative results of the reconstructed face images by different methods compared with the ground truth HR images in terms of PSNR values. From the tables, we can see that our method has the highest PSNR values across all testing face images.

 \begin{figure*}[!t]
 	\includegraphics[scale=.5]{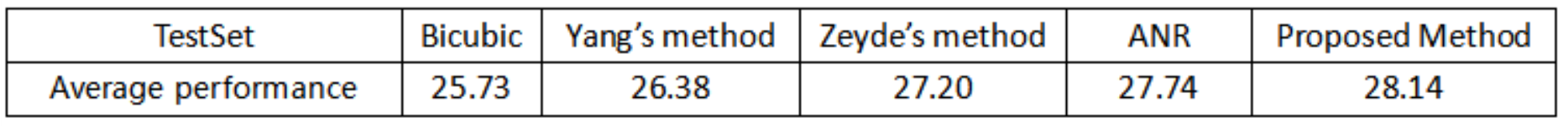}
	\caption*{{\bf Table 1.}
	The average results of PSNR (dB) by different methods on the AR dataset.}
 		\label{A}
 \end{figure*}

 \begin{figure*}[!t]
 	\includegraphics[scale=.5]{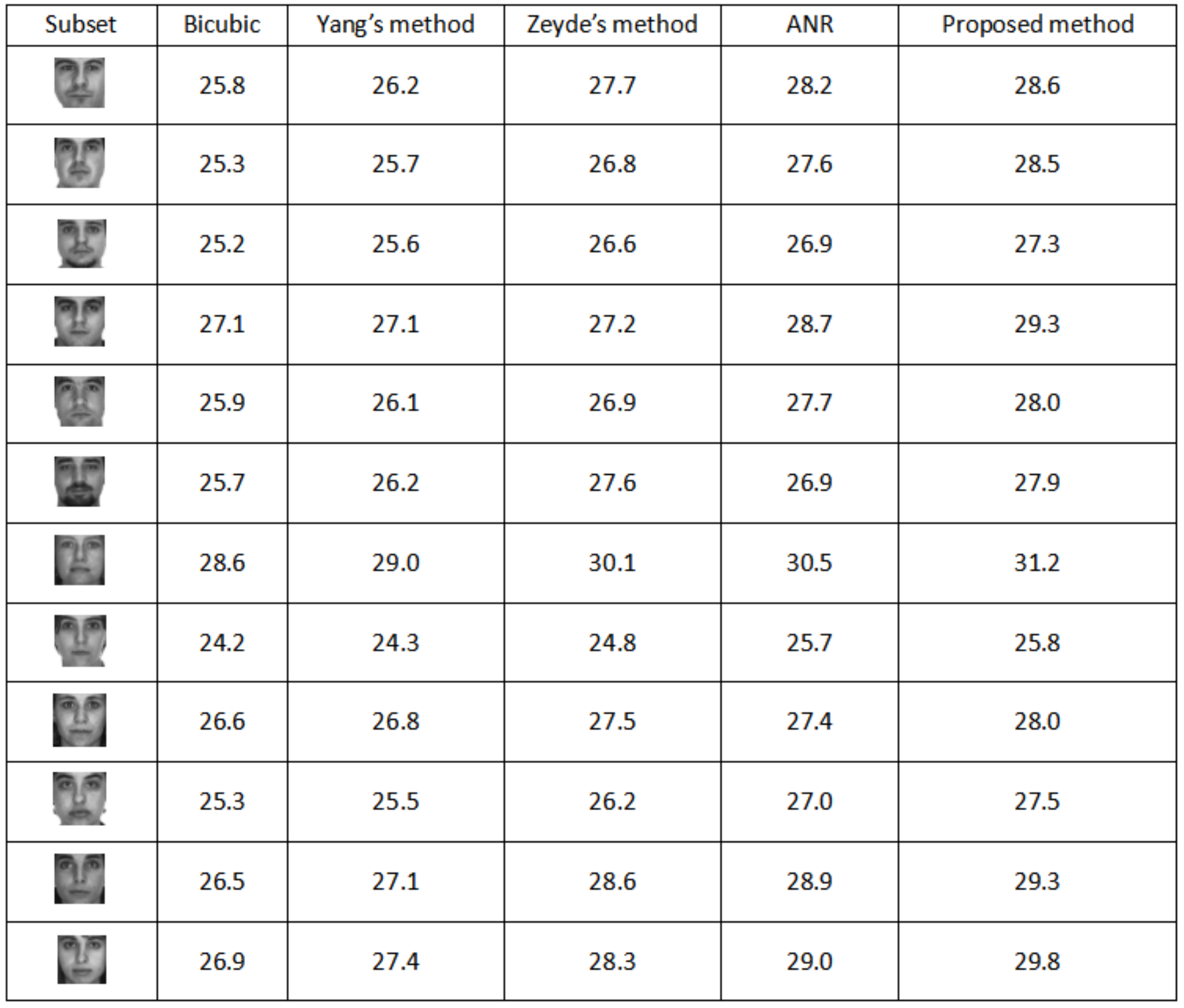}
	\caption*{{\bf Table 2.}
	Part of the experimental results on the AR dataset.}
 		\label{B}
 \end{figure*}

\section{Conclusion}
In this paper, we have proposed a novel weighted patches regression method to boost robust face hallucination performance. Instead of regularizing all facial patches equally, our proposed method discriminated different patches via an AdaBoost training procedure. In this way, the AdaBoost process can make us focus more on the facial regions that are more difficult to be correctly captured. Experimental results on two benchmark datasets demonstrate its superiority over state-of-the art methods in terms of PSNR values and visual quality.

\section*{Acknowledgments}
D.Z. is grateful for funding by the National Natural Science Foundation of China (Grant No.11401499), the Natural Science Foundation of Fujian Province of China (Grant No.2015J05016). Z.Z. is supported by National Natural Science Foundation of China (Grant No.61402389) and the Fundamental Research Funds for the Central Universities in China (No.20720160073).



\begin{thebibliography}{19}
\expandafter\ifx\csname natexlab\endcsname\relax\def\natexlab#1{#1}\fi
\providecommand{\url}[1]{\texttt{#1}}
\providecommand{\href}[2]{#2}
\providecommand{\path}[1]{#1}
\providecommand{\DOIprefix}{doi:}
\providecommand{\ArXivprefix}{arXiv:}
\providecommand{\URLprefix}{URL: }
\providecommand{\Pubmedprefix}{pmid:}
\providecommand{\doi}[1]{\href{http://dx.doi.org/#1}{\path{#1}}}
\providecommand{\Pubmed}[1]{\href{pmid:#1}{\path{#1}}}
\providecommand{\bibinfo}[2]{#2}
\ifx\xfnm\relax \def\xfnm[#1]{\unskip,\space#1}\fi
\bibitem[{Aharon et~al.(2006a)Aharon, Elad and Bruckstein}]{aharonk}
\bibinfo{author}{Aharon, M.}, \bibinfo{author}{Elad, M.},
  \bibinfo{author}{Bruckstein, A.}, \bibinfo{year}{2006}a.
\newblock \bibinfo{title}{The k-svd: An algorithm for designing of overcomplete
  dictionaries for sparse representation}.
\newblock \bibinfo{journal}{IEEE transactions on Signal Processing}
  \bibinfo{volume}{54}, \bibinfo{pages}{4311--4322}.
\bibitem[{Aharon et~al.(2006b)Aharon, Elad and Bruckstein}]{aharon2006img}
\bibinfo{author}{Aharon, M.}, \bibinfo{author}{Elad, M.},
  \bibinfo{author}{Bruckstein, A.}, \bibinfo{year}{2006}b.
\newblock \bibinfo{title}{K-svd: An algorithm for designing overcomplete
  dictionaries for sparse representation}.
\newblock \bibinfo{journal}{IEEE Transactions on Signal Processing}
  \bibinfo{volume}{54}, \bibinfo{pages}{4311--4322}.
\bibitem[{Chambolle(2004)}]{chambolle2004algorithm}
\bibinfo{author}{Chambolle, A.}, \bibinfo{year}{2004}.
\newblock \bibinfo{title}{An algorithm for total variation minimization and
  applications}.
\newblock \bibinfo{journal}{Journal of Mathematical imaging and vision}
  \bibinfo{volume}{20}, \bibinfo{pages}{89--97}.
\bibitem[{Chang et~al.(2010)Chang, Zhou, Han and Deng}]{chang2010face}
\bibinfo{author}{Chang, L.}, \bibinfo{author}{Zhou, M.}, \bibinfo{author}{Han,
  Y.}, \bibinfo{author}{Deng, X.}, \bibinfo{year}{2010}.
\newblock \bibinfo{title}{Face sketch synthesis via sparse representation}, in:
  \bibinfo{booktitle}{20th International Conference on Pattern Recognition},
  \bibinfo{organization}{IEEE}. pp. \bibinfo{pages}{2146--2149}.
\bibitem[{Drucker(1997)}]{drucker1997improving}
\bibinfo{author}{Drucker, H.}, \bibinfo{year}{1997}.
\newblock \bibinfo{title}{Improving regressors using boosting techniques}, in:
  \bibinfo{booktitle}{Machine Learning: Proceedings of the Fourteenth
  International Conference.}, pp. \bibinfo{pages}{479--485}.
\bibitem[{Freund and Schapire(1997)}]{freund1997decision}
\bibinfo{author}{Freund, Y.}, \bibinfo{author}{Schapire, R.E.},
  \bibinfo{year}{1997}.
\newblock \bibinfo{title}{A decision-theoretic generalization of on-line
  learning and an application to boosting}.
\newblock \bibinfo{journal}{Journal of Computer and System Sciences}
  \bibinfo{volume}{55}, \bibinfo{pages}{119--139}.
\bibitem[{Geman and Reynolds(1992)}]{geman1992constrained}
\bibinfo{author}{Geman, D.}, \bibinfo{author}{Reynolds, G.},
  \bibinfo{year}{1992}.
\newblock \bibinfo{title}{Constrained restoration and the recovery of
  discontinuities}.
\newblock \bibinfo{journal}{IEEE Transactions on Pattern Analysis \& Machine
  Intelligence} , \bibinfo{pages}{367--383}.
\bibitem[{Jung et~al.(2011)Jung, Jiao, Liu and Gong}]{jung2011position}
\bibinfo{author}{Jung, C.}, \bibinfo{author}{Jiao, L.}, \bibinfo{author}{Liu,
  B.}, \bibinfo{author}{Gong, M.}, \bibinfo{year}{2011}.
\newblock \bibinfo{title}{Position-patch based face hallucination using convex
  optimization}.
\newblock \bibinfo{journal}{IEEE Signal Processing Letters}
  \bibinfo{volume}{18}, \bibinfo{pages}{367--370}.
\bibitem[{Ma et~al.(2012)Ma, Luong, Philips, Song and Cui}]{ma2012sparse}
\bibinfo{author}{Ma, X.}, \bibinfo{author}{Luong, H.Q.},
  \bibinfo{author}{Philips, W.}, \bibinfo{author}{Song, H.},
  \bibinfo{author}{Cui, H.}, \bibinfo{year}{2012}.
\newblock \bibinfo{title}{Sparse representation and position prior based face
  hallucination upon classified over-complete dictionaries}.
\newblock \bibinfo{journal}{Signal processing} \bibinfo{volume}{92},
  \bibinfo{pages}{2066--2074}.
\bibitem[{Martinez(1998)}]{martinez1998ar}
\bibinfo{author}{Martinez, A.M.}, \bibinfo{year}{1998}.
\newblock \bibinfo{title}{The ar face database}.
\newblock \bibinfo{journal}{CVC Technical Report} .
\bibitem[{Mumford and Shah(1989)}]{mumford1989optimal}
\bibinfo{author}{Mumford, D.}, \bibinfo{author}{Shah, J.},
  \bibinfo{year}{1989}.
\newblock \bibinfo{title}{Optimal approximations by piecewise smooth functions
  and associated variational problems}.
\newblock \bibinfo{journal}{Communications on pure and applied mathematics}
  \bibinfo{volume}{42}, \bibinfo{pages}{577--685}.
\bibitem[{Phillips et~al.(2000)Phillips, Moon, Rizvi and
  Rauss}]{phillips2000feret}
\bibinfo{author}{Phillips, P.J.}, \bibinfo{author}{Moon, H.},
  \bibinfo{author}{Rizvi, S.A.}, \bibinfo{author}{Rauss, P.J.},
  \bibinfo{year}{2000}.
\newblock \bibinfo{title}{The feret evaluation methodology for face-recognition
  algorithms}.
\newblock \bibinfo{journal}{IEEE Transactions on Pattern Analysis and Machine
  Intelligence} \bibinfo{volume}{22}, \bibinfo{pages}{1090--1104}.
\bibitem[{Rudin et~al.(1992)Rudin, Osher and Fatemi}]{rudin1992nonlinear}
\bibinfo{author}{Rudin, L.I.}, \bibinfo{author}{Osher, S.},
  \bibinfo{author}{Fatemi, E.}, \bibinfo{year}{1992}.
\newblock \bibinfo{title}{Nonlinear total variation based noise removal
  algorithms}.
\newblock \bibinfo{journal}{Physica D: Nonlinear Phenomena}
  \bibinfo{volume}{60}, \bibinfo{pages}{259--268}.
\bibitem[{Timofte et~al.(2013)Timofte, Smet and Gool}]{timofte2013anchored}
\bibinfo{author}{Timofte, R.}, \bibinfo{author}{Smet, V.},
  \bibinfo{author}{Gool, L.}, \bibinfo{year}{2013}.
\newblock \bibinfo{title}{Anchored neighborhood regression for fast
  example-based super-resolution}, in: \bibinfo{booktitle}{Proceedings of the
  IEEE International Conference on Computer Vision}, pp.
  \bibinfo{pages}{1920--1927}.
\bibitem[{Timofte and Van~Gool(2014)}]{timofte2014adaptive}
\bibinfo{author}{Timofte, R.}, \bibinfo{author}{Van~Gool, L.},
  \bibinfo{year}{2014}.
\newblock \bibinfo{title}{Adaptive and weighted collaborative representations
  for image classification}.
\newblock \bibinfo{journal}{Pattern Recognition Letters} \bibinfo{volume}{43},
  \bibinfo{pages}{127--135}.
\bibitem[{Wang et~al.(2004)Wang, Bovik, Sheikh and Simoncelli}]{wang2004image}
\bibinfo{author}{Wang, Z.}, \bibinfo{author}{Bovik, A.C.},
  \bibinfo{author}{Sheikh, H.R.}, \bibinfo{author}{Simoncelli, E.P.},
  \bibinfo{year}{2004}.
\newblock \bibinfo{title}{Image quality assessment: from error visibility to
  structural similarity}.
\newblock \bibinfo{journal}{IEEE Transactions on Image Processing}
  \bibinfo{volume}{13}, \bibinfo{pages}{600--612}.
\bibitem[{Yang et~al.(2010)Yang, Wright, Huang and Ma}]{yang2010image}
\bibinfo{author}{Yang, J.}, \bibinfo{author}{Wright, J.},
  \bibinfo{author}{Huang, T.S.}, \bibinfo{author}{Ma, Y.},
  \bibinfo{year}{2010}.
\newblock \bibinfo{title}{Image super-resolution via sparse representation}.
\newblock \bibinfo{journal}{IEEE Transactions on Image Processing}
  \bibinfo{volume}{19}, \bibinfo{pages}{2861--2873}.
\bibitem[{Zeyde et~al.(2010)Zeyde, Elad and Protter}]{zeyde2010single}
\bibinfo{author}{Zeyde, R.}, \bibinfo{author}{Elad, M.},
  \bibinfo{author}{Protter, M.}, \bibinfo{year}{2010}.
\newblock \bibinfo{title}{On single image scale-up using
  sparse-representations}, in: \bibinfo{booktitle}{Curves and Surfaces}.
  \bibinfo{publisher}{Springer}, pp. \bibinfo{pages}{711--730}.
\bibitem[{Zhang et~al.(2012)Zhang, Zhao, Xiong, Ma and Zhao}]{zhang2012image}
\bibinfo{author}{Zhang, J.}, \bibinfo{author}{Zhao, C.},
  \bibinfo{author}{Xiong, R.}, \bibinfo{author}{Ma, S.}, \bibinfo{author}{Zhao,
  D.}, \bibinfo{year}{2012}.
\newblock \bibinfo{title}{Image super-resolution via dual-dictionary learning
  and sparse representation}, in: \bibinfo{booktitle}{2012 IEEE International
  Symposium on Circuits and Systems (ISCAS)}, \bibinfo{organization}{IEEE}. pp.
  \bibinfo{pages}{1688--1691}.

\end{thebibliography}

\end{document}